\documentclass{article}

\usepackage{microtype}
\usepackage{graphicx}
\usepackage{subfigure}
\usepackage{verbatim}
\usepackage{booktabs} 

\usepackage{hyperref}

\usepackage[accepted]{icml2021}

\icmltitlerunning{Reinforcement Learning for Assignment Problem with Time Constraints}

\begin{document}
\nocite{*}
\twocolumn[
\icmltitle{Reinforcement Learning for Assignment Problem with Time Constraints}



\icmlsetsymbol{equal}{*}

\begin{icmlauthorlist}
\icmlauthor{Sharmin Pathan}{equal,to}
\icmlauthor{Vyom Shrivastava}{equal,to}
\end{icmlauthorlist}

\icmlaffiliation{to}{University of Georgia, USA}

\icmlcorrespondingauthor{Sharmin Pathan}{sharmin.pathan07@gmail.com}
\icmlcorrespondingauthor{Vyom Shrivastava}{vyomshrivastava2@gmail.com}

\icmlkeywords{Reinforcement Learning, Combinatorial Optimization, Assignment Problem}

\vskip 0.3in
]\hyphenpenalty=10000



\printAffiliationsAndNotice{\icmlEqualContribution} 

\begin{abstract}
We present an end-to-end framework for the Assignment Problem with multiple tasks mapped to a group of workers, using reinforcement learning while preserving many constraints. Tasks and workers have time constraints and there is a cost associated with assigning a worker to a task. Each worker can perform multiple tasks until it exhausts its allowed time units (capacity). We train a reinforcement learning agent to find near optimal solutions to the problem by minimizing total cost associated with the assignments while maintaining hard constraints. We use proximal policy optimization to optimize model parameters. The model generates a sequence of actions in real-time which correspond to task assignment to workers, without having to retrain for changes in the dynamic state of the environment. In our problem setting reward is computed as negative of the assignment cost. We also demonstrate our results on bin packing and capacitated vehicle routing problem, using the same framework. Our results outperform Google OR-Tools using MIP and CP-SAT solvers with large problem instances, in terms of solution quality and computation time.
\end{abstract}

\section{Introduction}
The term 'Neural Combinatorial Optimization' proposed by \cite{bello} et al. is a framework to tackle combinatorial optimization problems using neural networks and reinforcement learning. This framework solves Traveling Salesman Problem (TSP) with up to 100 customer nodes on a Euclidean graph while achieving near optimal solutions. TSP requires finding the shortest route connecting all customer nodes and returning to the starting point. A few other combinatorial optimization problems like the vehicle routing problem (VRP), bin packing, and assignment problem follow a similar framework. Vehicle Routing Problem is a variant of Traveling Salesman Problem, with the use of multiple vehicles and several other constraints like capacity associated with different vehicles or time-windows within which a customer node needs to serviced. Bin packing problems focus on packing a set of items with varying sizes into fixed capacity containers. Every item has a value and dimension associated with it, and the goal is to fit maximum items into the available containers for maximum capacity utilization. The customer nodes in TSP/VRP become containers in bin-packing problem, and workers in the assignment problem. The assignment problem requires a group of workers to perform a certain set of tasks. These tasks have time constraints and costs associated with them against each worker, while the workers have fixed time unit limitations. The complexity of these problems comes from dynamic nature of the environment rather than intractability of computing the optimal solution. The use of neural networks and reinforcement learning becomes a compelling choice to work with problems of this nature, as the agent can learn by interacting with the environment by gathering rewards. The performance of reinforcement learning to solve combinatorial optimization problems is competitive, while not requiring much of the domain knowledge. \cite{nazari} et al. solve both static and online versions of vehicle routing problem by using actor-critic methods for reinforcement learning for 10, 20, 50, and 100 VRP instances with customer locations on a Euclidean graph and demands as the static and dynamic features of the input respectively, with negative distance as the reward. Critic estimates the value function and actor updates policy distribution in the direction suggested by critic (with policy gradients). Basically, an actor decides what action to take and critic evaluates how good the action was and how the actor should adjust. Both actor and critic are parameterized by neural networks.
\newline
\newline
\cite{schulman} et al. proposed Proximal Policy Optimization (PPO), a policy gradient method for reinforcement learning which has some benefits of trust region optimization, but are much simpler and have better sample complexity. It involves training on a small batch from experiences interacting with the environment and updating the policy while ensuring the updates don't deviate much from previous policy. A new batch of experiences is used for every update. In our framework, we use PPO for policy optimization in conjunction with actor-critic methods. The actor is trained to study environment state inputs characterized by 
effort required for a task and available time units with a worker. Critic estimates the value function based on costs measured by the cumulative cost associated with task-worker combination. The reinforcement learning agent learns this policy to produce solutions with minimal workers, thereby reducing costs. With every action taken, the policy compares it with another action to compute advantage of one action over another for the given environment state.
\newline
\newline
We propose a framework to solve the Assignment Problem with fixed time constraints using reinforcement learning to plan most cost-effective task assignments to worker groups. We also experiment with bin-packing to maximize available container utilization increasing associated value, and capacitated vehicle routing problem to plan multi-stop routes with least number of vehicles and shortest path, to eventually reduce transportation costs from geographically distributed customer locations. This framework has the potential to be applied more generally to combinatorial optimization problems. We formulate the problem as Markov Decision Process (MDP) for modeling a series of actions for decision making given the state of the environment. The environment state is characterized by the static and dynamic elements, and time constraints within which the nodes (workers/containers/customers) need to be serviced. The environment is continuously evolving until a termination condition is met. Selecting what node to service next is the action to be taken by the agent given current state of the environment. With every action taken, the agent accumulates a reward, with the goal of finding the most cost-effective value-driven solution.
\newline
\newline
With traditional heuristic approaches or when modeling on static instance-specific environments, a policy needs to be trained for every instance separately. Changes to the state would require to build solutions from scratch. For the assignment problem, we model on a dynamic environment where the task efforts and available worker time units change over time. Every task has a worker eligibility based on what type of worker can service the task. The trained policy can thus perform well on instances sampled from the same distribution. It can accommodate changes to the dynamic elements and can automatically adapt to the solution. The policy suggests best possible worker-task combinations by finding a solution with minimal cost and least number of workers. A worker can accept tasks until it runs out of available time units and other constraints. The process continues until there are no more tasks to be performed.

\begin{figure*}[ht]
\vskip 0.2in
  \centering
  \fbox{ \includegraphics[scale=0.4]{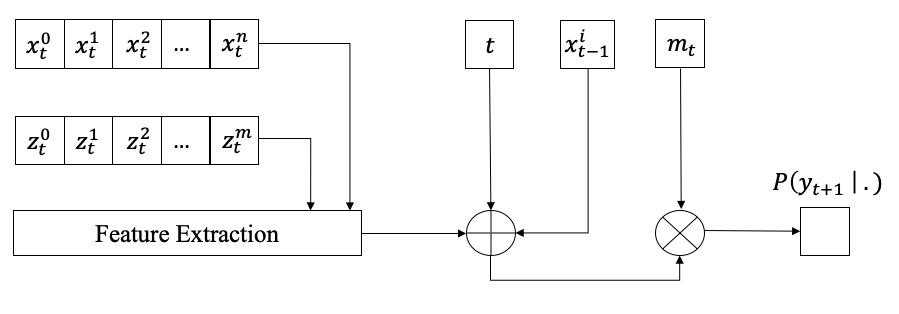} \rule[-.5cm]{0cm}{0cm} \rule[-.5cm]{0cm}{0cm}} 
  \caption{A schematic representation of our proposed network architecture. We extract features from the dynamic elements of environment state, concatenate it with current time (${t}$) and last completed task node (${x^i_{t-1}}$), and finally mask (${m_t}$) the non-serviceable customer nodes. \newline}
  \label{network}
  \vskip -0.2in
\end{figure*}

\section{Problem Formulation}
We focus on the Assignment Problem as our main design and use bin packing and capacitated vehicle routing as an extension to our experiments. The objective is to maximize rewards and using minimum workers / bins / vehicles. In the real world setting, it is more important to reduce total number of worker / bins / vehicles used, as these come with fixed capacities at fixed costs.

\subsection{Assignment Problem}
We define an environment state-action pair for the assignment problem with time constraints. The environment state is dynamically evolving with every action taken. In our problem setting, we define our environment state by a set of input features for every task ${X = \{x^i, i = 1, 2, .., n\}}$, and for every worker ${Z = \{z^j, j = 1, 2, .., m\}}$. These features are characterized by dynamic elements of the inputs over time, which is the effort (time units) for selected task and available time units with a worker. Every time a task ${x^i}$ is completed by worker ${z^j}$, ${x^i}$ effort is set to ${0}$, and ${z^j}$ available time units are updated. The environment state keeps evolving until the termination condition is met (all tasks are completed). For a given task, the agent selects a worker node eligible to service from a pool of different workers available.
\newline
\newline
We allow a worker to perform multiple tasks. Once a worker runs out of available time units or cannot service additional tasks due to other constraints, it is skipped for further tasks. The available set of actions at ${t=0}$ with ${m}$ worker nodes are ${Y = \{y^k, k = 1, 2, .., m\}}$. Not every worker is assigned a task if a subset of workers can finish off all tasks. The agent keeps producing a sequence of actions for evolving states until the termination condition is satisfied. We define the termination condition as, when all tasks have been completed. ${cost_t}$ is the sum of cost over time.
\newline
\newline
Once a worker ${z^j}$ finishes task ${x^i}$,

\[ cost_t = calculate\_cost(x^i_t, z^j_t) \]
\[ z^j_{t+1} = z^j_t - x^i_t \]
\[ x^i_{t+1} = 0 \]

We use a masking scheme that masks all worker nodes not eligible to service current set of tasks. These masked nodes include $(i)$ worker nodes with zero available time units $(z^j_t = 0)$ and $(ii)$ worker nodes with available time units less than minimum value of current task efforts $(X)$. The $calculate\_cost(task, worker)$ gives cost associated with the worker $z^j$ performing task $x^i$. We do not explicitly provide cost values to the model as input. Cost values are provided in reward formulation for the model to learn from. For the reward setting, we use negative of cost value and number of workers as the total reward. This eliminates the need to explicitly provide a cost matrix as input to the model. This reward works as feedback to the agent. The agent finds an optimal policy by maximizing rewards while satisfying problem constraints. The optimal policy  $\pi^*$ will generate the optimal solution with probability 1. Our goal is to make $\pi$ as close to $\pi^*$ as possible. 

\subsection{Bin Packing}
The bin packing environment state is defined by a collection of $n$ items of varying weights $X = {\{x^i, i = 1, 2, .., n\}}$, and bins with capacities $Z = {\{z^j, j = 1, 2, .., m\}}$. The problem is to fit maximum items into minimum bins while maximizing total value, with $m$ total available bins. Value is computed in the reward formulation. Bins that run out of capacity are masked. The available set of actions at ${t=0}$ with ${m}$ bins are ${Y = \{y^k, i = 1, 2, .., m\}}$. The environment is evolving with the sequence of actions taken by the agent until the termination condition is met. 

\subsection{Capacitated Vehicle Routing}
In the capacitated vehicle routing problem setting, our input features are characterized by customer demands $X = {\{x^i, i = 1, 2, .., n\}}$, and vehicles with capacities $Z = {\{z^j, i = 1, 2, .., m\}}$. The problem setting is similar to assignment problem and bin packing, to service maximum demands using minimum vehicles. Objective is to maximize reward which is the negative of distance between geographically distributed customer nodes. Vehicles that run out of capacity are masked. The available set of actions at ${t=0}$ with ${m}$ vehicles are ${Y = \{y^k, k = 1, 2, .., m\}}$. The environment is evolving with the sequence of actions taken by the agent until the termination condition is met. 

\section{Network architecture}
In our framework, we use proximal policy optimization (PPO) for policy optimization in conjunction with actor-critic methods. Both actor and critic are parameterized by neural networks. The advantage function basically decides how much better off it is to take a specific action. In our assignment problem, it decides among a choice of available worker nodes. We transform current reward with the future reward along with a discount factor $\gamma$ = 0.99 and maintain an experience buffer of length 1000. We set the learning rate to 1e-4 with a decay of 0.001. We lower the learning rate as the problem size grows. Lower learning rate greatly stabilizes training. Loss clipping ($\epsilon$) is set to 0.2 as the tolerance for policy updates for how much we are willing to deviate from the previous policy. Using this we ensure the policy lies within the trust region, where the local approximations of the policy are accurate. With reward from current state as ${r_t}$, value function from current state as ${V(X_t)}$, and value function from future state as ${V(X_{t+1})}$, the Generalized Advantage Estimation (GAE) for our advantage function is,

\[ GAE = r_t + \gamma \cdot V(X_{t+1}) - V({X_t}) \]

Let ${\pi_{new}}$ be the new policy, ${\pi_{old}}$ the old policy. Loss for the actor network is computed as follows:

\[ratio = [\log{\pi_{new}} - \log{\pi_{old}}]\cdot \exp() \]
\[ r\_gae = (ratio \cdot GAE) \]
\[ l\_gae = loss\_clipping(ratio, 1-\epsilon, 1+\epsilon) \cdot GAE) \]
\[actor\_loss = 	min(r\_gae, l\_gae) \]

\begin{table*} [t]
  \caption{A comparison of costs, and solution times using Google OR-Tools and our framework for AP10, AP20, AP30, AP40, and AP50 instances. Solution time is in minutes. Solution time for our framework is recorded after training the model.  \newline}
  \label{sample-table}
  \centering
  \begin{tabular}{llllll}
    \toprule
    & & \multicolumn{2}{c}{Google OR-Tools} & \multicolumn{2}{c}{Our Framework}                 \\
    \cmidrule(r){3-4}
    \cmidrule(lr){5-6}
    AP Instance & Workers assigned & Solution time    & Cost    & Solution time & Cost \\
    \midrule
    \multicolumn{1}{c}{AP 10} & \multicolumn{1}{r}{3} & \multicolumn{1}{r}{0.000169}  & \multicolumn{1}{r}{1240} & \multicolumn{1}{r}{0.000264} & \multicolumn{1}{r}{1240}    \\
    \multicolumn{1}{c}{AP 20} & \multicolumn{1}{r}{6} & \multicolumn{1}{r}{0.400000}  & \multicolumn{1}{r}{1582} & \multicolumn{1}{r}{0.000522} & \multicolumn{1}{r}{1582}       \\
    \multicolumn{1}{c}{AP 30} & \multicolumn{1}{r}{11} & \multicolumn{1}{r}{11.250000} & \multicolumn{1}{r}{2264} & \multicolumn{1}{r}{0.001090} & \multicolumn{1}{r}{2142}   \\
    \multicolumn{1}{c}{AP 40} & \multicolumn{1}{r}{14} & \multicolumn{1}{r}{67.230000} & \multicolumn{1}{r}{3528} & \multicolumn{1}{r}{0.001110} & \multicolumn{1}{r}{3528} \\
    \multicolumn{1}{c}{AP 50} & \multicolumn{1}{r}{17} & \multicolumn{1}{r}{103.000000} & \multicolumn{1}{r}{9850} & \multicolumn{1}{r}{0.001580} & \multicolumn{1}{r}{8800}  \\
    \bottomrule
  \end{tabular}
  \label{or-comparison}
  \vskip 0.2in
\end{table*}

\begin{table*}[!htb]
  \caption{A comparison of assignment costs, and solution times using Google OR-Tools and our framework after updating time efforts of 5 tasks in AP30 instance. Solution time is in minutes. Updated task efforts doesn't require retraining our RL framework.  \newline}
  \label{sample-table}
  \centering
  \begin{tabular}{llllll}
    \toprule
    & & \multicolumn{2}{c}{Google OR-Tools} & \multicolumn{2}{c}{Our Framework}                 \\
    \cmidrule(r){3-4}
    \cmidrule(lr){5-6}
    AP Instance & Workers assigned & Solution time    & Cost    & Solution time & Cost \\
    \midrule
    \multicolumn{1}{c}{updated AP 30} & \multicolumn{1}{r}{10} & \multicolumn{1}{r}{12.327} & \multicolumn{1}{r}{2752} & \multicolumn{1}{r}{0.001090} & \multicolumn{1}{r}{2592}   \\
    \bottomrule
  \end{tabular}
  \label{updated30}
\end{table*}

Figure \ref{network} illustrates our proposed model. We use a simple neural network architecture for the actor and critic. The actor and critic architectures are identical, input to which is the environment state defined by task efforts and worker's available time units (capacity), current time and last task completed. The state input is then fed to a one dimensional convolution with 128 filters followed by a dense layer with 128 units. Current time and last performed task inputs are followed by two dense layers with 128 units. Final layer to the actor network is a dense layer with total\_workers units and softmax activation. The final layer to actor network further goes through a masking layer which masks $(i)$ worker nodes with zero available time units (capacity) and, $(ii)$ worker nodes with available time units (capacity) less than minimum value of current task efforts. We tried setting penalties for when the model makes a wrong decision and selects an invalid node, but masking facilitated faster training and convergence as the model has less options to learn from. When training, the actor has two additional inputs, another action and advantage to compare with the current prediction. Final layer to critic network is a 1 unit dense layer for the value function. The actor network uses proximal policy optimization loss function and critic uses mean squared error. We train the actor-critic networks for 20 epochs on every episode with a batch size of 256. By adding current time and last task completed as one of the inputs to actor and critic networks, we eliminate the need for an RNN LSTM decoder to encode sequence information.

\section{Experiments}
A typical assignment problem is to find an optimal assignment of tasks to a group of workers that minimizes the cost of these assignments. Several variants of the problem with additional constraints have been studied which include assignments across different teams of workers with a limit on number of tasks a team can complete, or a certain group of workers is eligible to undertake a task. Another form is when tasks have associated time constraints. We experiment with worker eligibilities across tasks with time constraints. We allow workers to undertake multiple tasks until it runs out of available time units (capacity). We make an initial split of worker nodes according to eligibility across tasks. All workers that are eligible to undertake a task are clustered together and the framework finds optimal assignments for that cluster. This now repeats fresh for the next cluster which has a different vehicle eligibility. We compare our results against Google OR-Tools. \cite{ortools} uses MIP and CP-SAT solver variants to solve the problem and we compare results against the best versions.
\newline
\begin{table}[t]
  \caption{Assignment costs on AP10 instance by increasing time efforts of every task by 5 units using Google OR-Tools and our pretrained policy on AP10 instance.  \newline}
  \label{sample-table}
  \centering
  \begin{tabular}{lll}
    \toprule
    & \multicolumn{2}{c}{Assignment Cost}                 \\
    \cmidrule(r){2-3}
    Tasks updated & Google OR-Tools    & Our Framework \\
    \midrule
    \multicolumn{1}{c}{1} & \multicolumn{1}{c}{1288} & \multicolumn{1}{c}{\textbf{1288}} \\
    \multicolumn{1}{c}{2} & \multicolumn{1}{c}{1316} & \multicolumn{1}{c}{1338} \\
    \multicolumn{1}{c}{3} & \multicolumn{1}{c}{1350} & \multicolumn{1}{c}{\textbf{1344}} \\
    \multicolumn{1}{c}{4} & \multicolumn{1}{c}{1350} & \multicolumn{1}{c}{\textbf{1350}} \\
    \multicolumn{1}{c}{5} & \multicolumn{1}{c}{1372} & \multicolumn{1}{c}{\textbf{1332}} \\
    \multicolumn{1}{c}{6} & \multicolumn{1}{c}{1400} & \multicolumn{1}{c}{\textbf{1392}} \\
    \multicolumn{1}{c}{7} & \multicolumn{1}{c}{1400} & \multicolumn{1}{c}{\textbf{1400}} \\
    \multicolumn{1}{c}{8} & \multicolumn{1}{c}{1420} & \multicolumn{1}{c}{1437} \\
    \multicolumn{1}{c}{9} & \multicolumn{1}{c}{1461} & \multicolumn{1}{c}{\textbf{1457}} \\
    \multicolumn{1}{c}{10} & \multicolumn{1}{c}{1494} & \multicolumn{1}{c}{1520} \\
    \bottomrule
  \end{tabular}
  \label{updated10}
  \end{table}

We use randomly generated tasks with time efforts and workers with capacities. We generate samples with 10, 20, 30, 40, 50 tasks, where the available worker nodes equal number of tasks with additional 2 worker nodes. The worker capacities are set to 15, and task time efforts are randomly generated with a cap at 15. We remove a worker node when it runs out of capacity or can no longer service additional tasks due to other constraints. We do not explicitly provide cost matrix to our framework, this is part of the reward formulation for the model to learn. Before starting with a solution, we give priority to tasks with time efforts equal to worker capacity, leaving less combinations for the model to try. So now the model has to find assignments only for the remaining workers and tasks. Table \ref{or-comparison} illustrates the solution times and resulting assignment costs for our experiments
\newline
\newline
Next, we modify task efforts of randomly selected 5 tasks in our AP30 instance and make predictions using Google OR-Tools and our reinforcement leaning framework. With Google's OR Tools and other traditional heuristic approaches, even modifying demand of a single customer node requires to build solutions from scratch. As shown in Table \ref{updated30}, OR Tools require another 12 minutes to build a solution for the same task nodes. Task time effort being the dynamic element in our framework's environment state, the previously trained policy can perform well on instances sampled from the same distribution. It accommodates these updates to task efforts and automatically adapts to a solution without the need to retrain the policy. Similarly, new tasks can replace already serviced tasks in our environment state and the previously trained policy can still make optimal predictions.
\newline
\newline
We increase time efforts of task nodes by 5 units on AP10 instance from Table \ref{updated30} and record total cost using Google OR-Tools. We then use our pretrained policy and make predictions on these updated task efforts. As shown in Table 3, we increase time effort of one task node by 5 units per iteration. OR-Tools converges to an optimal solution and we use it as a baseline to monitor if our pretrained policy makes optimal predictions with updates to the dynamic elements. There is no definite pattern as to how often a pretrained policy would make optimal predictions, but it does most of the times. Also, for larger AP instances where OR-tools fail to achieve near-optimal solutions, our framework with a pretrained policy achieves better solutions as experimented with AP 30 instance in Table \ref{updated30}.
\newline
\newline
Similar experiments from Table \ref{updated30} follow for bin packing problem and vehicle routing problem as illustrated in Table \ref{bin_updated10} and Table \ref{vrp_updated10} respectively. The results show our framework can be generalized across different environment types of combinatorial optimization. The trained policy can thus perform well on instances sampled from the same distribution. It can accommodate changes to the dynamic elements and can automatically adapt to the solution.

\begin{table}[!h]
  \caption{Total value on BIN10 instance by increasing weights of every item by 5 units using Google's OR-Tools and our pretrained policy on BIN10 instance.  \newline}
  \label{sample-table}
  \centering
  \begin{tabular}{lll}
    \toprule
    & \multicolumn{2}{c}{Total Value}                 \\
    \cmidrule(r){2-3}
    Demands updated & OR-Tools    & Our Framework \\
    \midrule
    \multicolumn{1}{c}{1} & \multicolumn{1}{c}{395} & \multicolumn{1}{c}{\textbf{395}} \\
    \multicolumn{1}{c}{2} & \multicolumn{1}{c}{409} & \multicolumn{1}{c}{421} \\
    \multicolumn{1}{c}{3} & \multicolumn{1}{c}{451} & \multicolumn{1}{c}{\textbf{444}} \\
    \multicolumn{1}{c}{4} & \multicolumn{1}{c}{532} & \multicolumn{1}{c}{\textbf{516}} \\
    \multicolumn{1}{c}{5} & \multicolumn{1}{c}{612} & \multicolumn{1}{c}{620} \\
    \multicolumn{1}{c}{6} & \multicolumn{1}{c}{649} & \multicolumn{1}{c}{663} \\
    \multicolumn{1}{c}{7} & \multicolumn{1}{c}{815} & \multicolumn{1}{c}{\textbf{809}} \\
    \multicolumn{1}{c}{8} & \multicolumn{1}{c}{820} & \multicolumn{1}{c}{814} \\
    \multicolumn{1}{c}{9} & \multicolumn{1}{c}{834} & \multicolumn{1}{c}{\textbf{828}} \\
    \multicolumn{1}{c}{10} & \multicolumn{1}{c}{850} & \multicolumn{1}{c}{\textbf{845}} \\
    \bottomrule
  \end{tabular}
  \label{bin_updated10}
\end{table}

\begin{table}[!h]
  \caption{Distance traveled on VRP10 instance by increasing demands of every customer by 5 units using Google's OR-Tools and our pretrained policy on VRP10 instance.  \newline}
  \label{sample-table}
  \centering
  \begin{tabular}{lll}
    \toprule
    & \multicolumn{2}{c}{Total Distance}                 \\
    \cmidrule(r){2-3}
    Demands updated & OR-Tools    & Our Framework \\
    \midrule
    \multicolumn{1}{c}{1} & \multicolumn{1}{c}{4496} & \multicolumn{1}{c}{4816} \\
    \multicolumn{1}{c}{2} & \multicolumn{1}{c}{5112} & \multicolumn{1}{c}{\textbf{5112}} \\
    \multicolumn{1}{c}{3} & \multicolumn{1}{c}{5240} & \multicolumn{1}{c}{\textbf{5240}} \\
    \multicolumn{1}{c}{4} & \multicolumn{1}{c}{5240} & \multicolumn{1}{c}{5315} \\
    \multicolumn{1}{c}{5} & \multicolumn{1}{c}{5432} & \multicolumn{1}{c}{\textbf{5432}} \\
    \multicolumn{1}{c}{6} & \multicolumn{1}{c}{5432} & \multicolumn{1}{c}{\textbf{5432}} \\
    \multicolumn{1}{c}{7} & \multicolumn{1}{c}{5432} & \multicolumn{1}{c}{5560} \\
    \multicolumn{1}{c}{8} & \multicolumn{1}{c}{5592} & \multicolumn{1}{c}{5820} \\
    \multicolumn{1}{c}{9} & \multicolumn{1}{c}{6344} & \multicolumn{1}{c}{\textbf{6344}} \\
    \multicolumn{1}{c}{10} & \multicolumn{1}{c}{6800} & \multicolumn{1}{c}{\textbf{6800}} \\
    \bottomrule
  \end{tabular}
  \label{vrp_updated10}
\end{table}

\section{Conclusion and Future Work}
Our proposed architecture can be extended to reuse of workers / bins / vehicles, and to work with more complex environments with several soft constraints like break intervals, driver shifts, and other variable costs such as fuel, in VRP. In our network design, using time as one of the inputs eliminates the need of adding sequential information to the model, thereby making the architecture much simpler. Also, it does not require an external cost/value matrix as this information is embedded in the reward function. Our proposed architecture can potentially be applied to various other combinatorial optimization problems. A real world and practical application would be managing ad-spaces to fill minimum slots and based on priority. Another application is to contract vehicles with third party providers. The model learns to adjust according to dynamic elements of the environment which eliminates need to retrain, if samples are generated from the same distribution. Masking helps with faster convergence and with larger problem sizes, our architecture provides competitive solution times.

\bibliography{example_paper}
\bibliographystyle{icml2021}

\end{document}